\documentclass[preprint,12pt]{elsarticle}
\usepackage{geometry}
\usepackage{lineno}

\usepackage{enumitem}
\usepackage{booktabs}
\usepackage{array}
\usepackage{longtable}
\usepackage{graphicx}
\graphicspath{{../}{./}}
\usepackage{hyperref}
\usepackage{xcolor}
\usepackage{pifont} 
\usepackage{url}
\usepackage{fancyhdr}
\usepackage{comment}
\usepackage{tikz}
\usepackage{orcidlink}
\usetikzlibrary{arrows.meta,positioning,shapes.geometric}
\usepackage{pgfplots}
\pgfplotsset{compat=1.18}

\usepackage{listings}
\usepackage{courier}
\renewcommand{\footrulewidth}{0pt}

\newcolumntype{L}[1]{>{\raggedright\arraybackslash}p{#1}}
\newcolumntype{R}[1]{>{\raggedright\arraybackslash}p{#1}}

\usepackage{amssymb}
\usepackage{amsmath}

\begin{document}

\fancyfoot[C]{\thepage}
\renewcommand{\footrulewidth}{0pt}

\section*{Methods Article}

\subsection*{Method title}
\noindent\textbf{A Reproducible Optimisation Protocol for Calibrating Prompt-Based Large Language Model Workflows in Evidence Synthesis}

\subsection*{Author}
\noindent\textbf{Teo Susnjak\orcidlink{0000-0001-9416-1435}}


\noindent School of Mathematical and Computational Sciences

\noindent Massey University, 

\noindent Auckland, New Zealand

\noindent Email: \textbf{T.Susnjak@massey.ac.nz}

\subsection*{Keywords}
\noindent\textbf{Evidence synthesis automation; Language model calibration; Prompt optimisation; Declarative programming; Systematic reviews; Reproducible workflows}

\section*{Abstract}
\noindent This methods article presents a reproducible calibration workflow for prompt-based large language models (LLMs) in structured evidence-synthesis tasks. The method separates the rules that define the scientific task from the mutable prompt harness that frames and applies them. It optimises that harness against labelled or reference examples and an explicit task metric, then preserves the calibrated workflow as an inspectable artefact with its specification, metric, settings, and evaluation traces. The example code instantiates the protocol with DSPy and GEPA tools, but the underlying logic can transfer to other prompt-optimisation frameworks that support structured task definitions, metric-guided search, and artefact reuse. Title and abstract screening is the worked validation case because it provides labelled benchmark data and clear evaluation metrics. The demonstrated workflow uses a smaller \textit{student} LLM for performing the scientific task execution and a larger \textit{reflection} LLM to steer the prompt optimisation process during calibration. This work shows compilation, artefact round-tripping, and how optimisation budget affects a smaller student model.

Key points:
\vspace{0.5em}
\begin{itemize}[leftmargin=4.2em]
  \item Separate \textit{what} the model must decide, the fixed rules, from \textit{how} it is prompted to decide, the mutable prompt instructions.
  \item Optimise the prompt automatically against labelled examples and a declared scoring function,
        rather than iterating manually.
  \item Package the calibrated prompt with its rules, metric, settings, traces, and evaluation logs for
        inspection and reuse.
\end{itemize}


\section*{Graphical abstract}

\vspace{0.75em}
\begin{center}
  \includegraphics[width=0.95\linewidth]{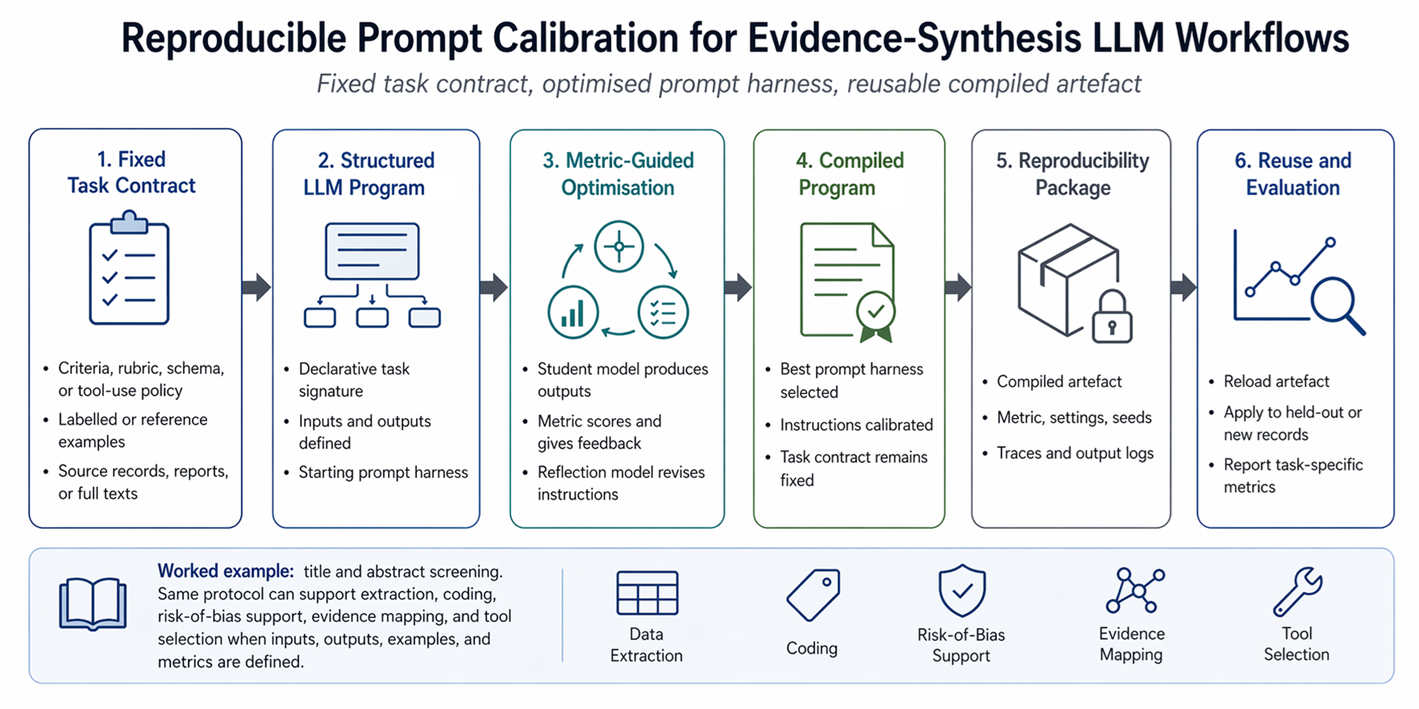}
\end{center}

\section*{Specifications table}
\begin{table}[h!]
\centering
\fontsize{10pt}{11pt}\selectfont
\renewcommand{\arraystretch}{1.2}
\begin{tabular}{L{0.32\linewidth} R{0.62\linewidth}}
\toprule
\textbf{Item} & \textbf{Description} \\
\midrule
Subject area & Computer Science; Information Science \\
More specific subject area & Evidence synthesis automation; research workflow automation; systematic review methods; language-model prompt optimisation; reproducible LLM workflows \\
Name of your method & Reproducible prompt calibration for structured evidence-synthesis LLM workflows \\
Name and reference of original method & DSPy declarative language-model programming \cite{khattab2023dspy,opsahl2024optimizing}; GEPA reflective prompt evolution \cite{agrawal2025gepa} \\
Resource availability & Companion Colab notebook, local test script, saved artefact, and validation summaries are provided with the manuscript materials. \\
\bottomrule
\end{tabular}
\end{table}

\section*{Background}
Large language models (LLMs) now offer a promising path for automating systematic literature reviews (SLRs) and meta analyses \cite{lieberum2025scoping}. Recent studies show that models can support screening \cite{Sujau2025Accelerating}, data extraction \cite{li2025levelautomationgoodenough}, quality assessment \cite{li2025automatedriskofbiasassessmentrandomized}, and synthesis tasks \cite{Susnjak2025Automating}. Currently, this promise has a practical limit. Performance depends not only on the task and model, but also on prompt wording, model family, output-generation settings (e.g., temperature and top-p parameters), and the manual prompt-tweaking choices made during iterative development. Studies show that prompt sensitivity varies across datasets and models. Larger models are often more robust, but pockets of variability persist even in stronger instruction-tuned systems \cite{zhuo-etal-2024-prosa,chatterjee-etal-2024-posix,Gooding2024}. Repeated prompting is methodologically important because LLM outputs are probabilistic and stability should be measured rather than assumed \cite{gallo2025repeatprompt}. These sensitivities matter most when researchers use smaller or locally hosted models because these LLMs tend to be cheaper and more privacy-preserving for sensitive domains. The consequence is that these models often need more calibration before they behave as reliable scientific instruments. In these cases, principled calibration can make LLM-based evidence-synthesis workflows more reliable, repeatable, and auditable.

  \subsection*{Prompt variability and the need for calibration}
Natural language has been framed as a new programming interface for LLMs \cite{kearns2023responsible}, but this analogy understates its ambiguity relative to the rigid syntax of conventional programming languages. Given that prompt performance is often contingent on numerous factors \cite{zhuo-etal-2024-prosa,chatterjee-etal-2024-posix}, \textit{prompt engineering} frequently turns into \textit{ad hoc} manual refinement. While studies report prompt-dependent variability on models of all sizes \cite{zhuo-etal-2024-prosa,Gooding2024} and brittleness in SLR automation  \cite{staudinger2024reproducibility,lieberum2025scoping}, they generally do not provide reusable calibration protocols for structured evidence-synthesis tasks that preserve repeatability and reproducibility. To that end, this article treats prompt optimisation as calibration of a scientific measurement instrument. The instrument is a workflow that maps a structured research object, such as a title, abstract, full text, table, rubric, or tool request, to a structured output or decision. 
Even though larger and frontier commercial models may need less calibration, they nonetheless require an auditable task definition and demonstrable validation metric that support confidence in prompt stability and accuracy. 

This methodological study makes three contributions. First, it formalises prompt optimisation as calibration of a mutable prompt harness around a fixed scientific task contract. Second, it shows how an executable metric can encode task requirements, output validity, and workflow trade-offs, using title and abstract screening as the worked validation case. Third, it defines a transparency pattern in which the compiled artefact, fixed task specification, metric, optimisation configuration, environment record, traces, and prediction logs are retained as inspectable artefacts. These contributions are demonstrated with DSPy, a declarative language model programming framework \cite{khattab2023dspy,opsahl2024optimizing}, and GEPA, a reflective prompt optimiser \cite{agrawal2025gepa}, but the protocol is intended to generalise to other prompt-optimisation frameworks with structured task definitions, executable metrics, and reusable compiled outputs. Thus, the same protocol can be implemented in other optimisation toolchains that support structured task definitions, executable metrics, metric-guided search, and reusable outputs \cite{pmlr-v293-spiess25a}.

\section*{Method details}
\subsection*{Overview}

The method requires four ingredients: a fixed task contract, which defines the scientific requirements; a mutable prompt harness, which defines how those requirements are framed and operationalised for the model; labelled or reference examples, which provide calibration and held-out evaluation cases; and an executable metric, which scores outputs against the task policy and provides feedback for optimisation. Together, these components define the calibration setting. The protocol performs bounded optimisation of the prompt harness against the metric and preserves the calibrated workflow as a reusable artefact.

In DSPy, the calibrated workflow is represented as a \textit{language model (LM) program}. This does not mean a conventional software program in the usual sense. It means a structured object that combines input fields, output fields, task instructions, model calls, and evaluation logic. In this article, program refers to this DSPy-style prompt-based workflow, namely, the object that GEPA calibrates and that can later be saved, reloaded, inspected, and reused.
This work also does not introduce a new prompt-optimisation algorithm. Instead, it contributes a \textit{reproducible methodological pattern} for applying existing prompt-optimisation frameworks to structured evidence-synthesis and scientific LLM workflows. The protocol demonstrates how to (1) fix the scientific task contract, (2) calibrate the prompt harness around that contract, (3) make the optimisation metric explicit, and (4) preserve the compiled program with enough provenance for inspection and reuse.

Title and abstract screening is a helpful use case because it is a common and well-defined evidence-synthesis task with labelled benchmark data and clear evaluation metrics. The same calibration pattern can, in principle, be adapted to other repeatable and structured evidence-synthesis tasks, including data extraction, study-characteristic coding, risk-of-bias support, evidence-domain mapping, and search-query generation, provided that the task has well-defined inputs, outputs, reference examples, and an executable metric. The protocol is less directly applicable to open-ended synthesis, exploratory hypothesis generation, or interpretive judgement tasks where no stable reference output or scoring policy can be defined. In those cases, the same packaging principles remain useful, but metric-guided prompt optimisation becomes harder to validate.
Again, the critical requirement is that inputs, outputs, reference examples, and task metrics can be clearly expressed. As a way of example, \ref{app:meta_analysis_extraction} gives an alternative schematic for a data-extraction adaptation. The accompanying Colab notebook\footnote{\texttt{MethodsX\_GEPA\_Tutorial.ipynb} accessible at \url{https://colab.research.google.com/drive/1gY_XzQ1h1EsK18bU136t2euiNWlo7sQC?usp=sharing}}, contains the complete runnable Python source code. Meanwhile, the article body here includes the key method-defining code snippets only while small example records and notebook-specific execution cells are confined to the Colab notebook to preserve readability.
 The presented protocol has four steps:

\begin{enumerate}[
    label=\arabic*.,
    leftmargin=4em,
    labelsep=0.6em
]
    \item define the task contract as a structured LM-program signature;
    \item codify the task standard using labelled or reference examples and an executable metric;
    \item compile the prompt harness with an optimisation framework under a fixed budget;
    \item save, reload, evaluate, and package the compiled artefact.
\end{enumerate}

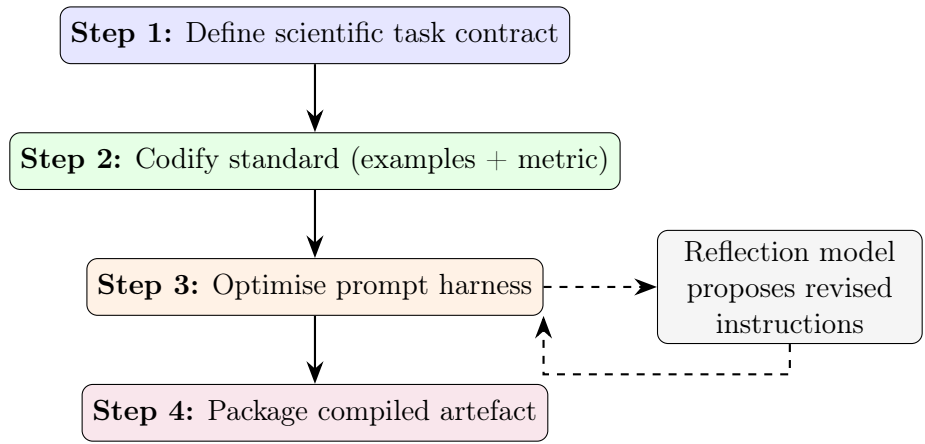
\begin{figure}[htbp]
\centering
\begin{tikzpicture}[
    node distance=0.9cm and 0.5cm,
    box/.style={draw, rounded corners, minimum width=4.8cm, minimum height=0.75cm,
                text centered, font=\small},
    arr/.style={-{Stealth[length=3mm]}, thick},
]
  \node[box, fill=blue!10]  (sig)   {\textbf{Step 1:} Define scientific task contract};
  \node[box, fill=green!10, below=of sig]  (std)   {\textbf{Step 2:} Codify standard (examples + metric)};
  \node[box, fill=orange!10, below=of std] (comp)  {\textbf{Step 3:} Optimise prompt harness};
  \node[box, fill=purple!10, below=of comp](pkg)   {\textbf{Step 4:} Package compiled artefact};

  \node[box, fill=gray!8, right=1.5cm of comp, minimum width=3.5cm, text width=3.2cm]
        (refl) {Reflection model proposes revised instructions};

  \draw[arr] (sig)  -- (std);
  \draw[arr] (std)  -- (comp);
  \draw[arr] (comp) -- (pkg);
  \draw[arr, dashed] (comp.east) -- (refl.west);
  \draw[arr, dashed] (refl.south) -- ++(0,-0.4) -| (comp.south east);
\end{tikzpicture}
\caption{Generic calibration workflow for structured prompt-based LLM programs. The title-and-abstract screening experiment is one instantiation of this workflow. Dashed arrows show the reflective optimisation loop inside GEPA.}
\label{fig:workflow}
\end{figure}

Figure~\ref{fig:workflow} summarises the four-step calibration workflow. Step~1 defines the scientific task contract, Step~2 turns that contract into labelled or reference examples and an executable metric, Step~3 optimises the prompt harness, and Step~4 packages the compiled artefact for inspection, reuse, and evaluation. Table~\ref{tab:fixed_mutable} clarifies what is meant by ``fixed'' and ``mutable'' in this workflow, while Table~\ref{tab:task_generality} shows how the same calibration pattern can be adapted and is generalisable beyond title-and-abstract screening. The fixed/mutable distinction is best understood as a three-layer structure. The first layer is the scientific task contract: the substantive rules that define the task, such as eligibility criteria, an extraction schema, a coding manual, a risk-of-bias rubric, reference labels, or a review policy. The second layer is the machine-readable interface: the input fields, output fields, label set, parsing rules, and validity constraints that define what the program must receive and return during a calibration run. These first two layers remain fixed. 
The third layer is the mutable prompt harness. This is the instructional wrapper around the fixed task contract and interface. It may include task framing, domain context, decision guidance, near-miss explanations, uncertainty handling, examples of acceptable checks, and trace-format guidance. This layer is where \textit{context engineering} occurs, and it is the only layer GEPA is permitted to revise and attempt to optimally frame for a target LLM. In this article, the phrase ``fixed task contract'' thus refers to the scientific task contract and the machine-readable interface together, unless otherwise specified.

\begin{table}[t]
\centering
\fontsize{10pt}{11pt}\selectfont
\begin{tabular}{p{0.24\linewidth}p{0.22\linewidth}p{0.44\linewidth}}
\toprule
\textbf{Layer} & \textbf{Fixed or mutable?} & \textbf{Examples} \\
\midrule
Scientific task contract & Fixed & Eligibility criteria, extraction schema, coding manual, reference labels, review policy \\
Machine-readable interface & Fixed within a calibration run & Input fields, output fields, label set, JSON or DSPy markers, parser, output-validity rules \\
Prompt harness & Mutable & Task wording, decision guidance, near-miss guidance, examples of checks, edge-case instructions \\
\bottomrule
\end{tabular}
\caption{Three-layer architecture separating the scientific task rules, the machine-readable interface, and the mutable prompt harness. During calibration, GEPA may revise the prompt harness but not the scientific task contract or the interface. Abstract screening is one instantiation of this pattern.}

\label{tab:fixed_mutable}
\end{table}

\begin{table}[htbp]
\centering
\fontsize{10pt}{11pt}\selectfont
\begin{tabular}{p{0.20\linewidth}p{0.26\linewidth}p{0.26\linewidth}p{0.20\linewidth}}
\toprule
\textbf{Task type} & \textbf{Fixed task contract} & \textbf{Mutable prompt harness} & \textbf{Possible metric} \\
\midrule
Title/abstract screening & Inclusion/exclusion criteria and labelled records & Screening instructions, uncertainty guidance, edge-case handling & Recall, precision, $F_1$, workload utility \\
Data extraction & Extraction schema and reference extractions & Extraction order, field-use guidance, evidence-span instructions & Exact or partial field match; evidence support score \\
Full-text eligibility & Eligibility criteria and adjudicated decisions & Clause-by-clause assessment procedure & Agreement with reviewer labels; false-exclusion penalty \\
Risk-of-bias support & Assessment rubric and expert ratings & Domain-specific rating instructions and justification format & Weighted agreement; class-pair cost matrix \\
Study-characteristic coding & Coding manual and labelled examples & Category boundary guidance and examples & Macro-$F_1$; class-balanced agreement \\
Evidence-domain mapping & Domain taxonomy and validated mappings & Taxonomy application procedure & Multiclass agreement; confusion-cost utility \\
Search-query generation & Search objective and known relevant studies & Query-construction and expansion instructions & Recall of known studies; query precision proxy \\
Tool selection or workflow routing & Available tools, routing policy, and reference decisions & Tool-choice instructions and failure-handling rules & Tool-choice accuracy; cost-weighted routing utility \\
\bottomrule
\end{tabular}
\caption{Examples of repeatable evidence-synthesis and research-support tasks that can be expressed using the same calibration protocol. The present paper validates the protocol using title and abstract screening, but the methodological pattern applies whenever a task can be specified through structured inputs, outputs, reference examples, and an executable metric.}
\label{tab:task_generality}
\end{table}

\subsection*{Materials and resources}

The method shown uses Python, a prompt-optimisation framework, labelled or reference examples, a \textit{student} language model for task execution, and a \textit{reflection} language model for prompt revision. 
The student model is the LLM of interest and is the target model which is to be deployed after the prompt calibration. In settings with cost, privacy, or governance constraints, this will usually be the smaller, less expensive or locally hosted model. The reflection model can be more capable and expensive because it is used only during compilation, and its role is significantly less token-intensive. It inspects failures, reads metric feedback, and proposes revised instructions for the student program. This separation allows researchers to   economically balance large LLM reasoning capacity during calibration without requiring the same model for routine task execution. Recent literature shows that calibrated smaller models can, in some settings, even approach larger-model performance at substantially lower cost \cite{zhao2025egoprompt,agrawal2025gepa}.

Temperature settings control the two roles differently. The student model is run with \texttt{temperature =0.0} to reduce avoidable variation during evaluation and execution of the scientifc task. The reflection model is run with a higher temperature, typically around \texttt{temperature=1.0}, because reflective prompt optimisation benefits from creative and diverse candidate revisions during search. The Colab notebook uses example provider strings only; comparable smaller models can be substituted for the student and stronger models can likewise be substituted for reflection\footnote{API credentials must be supplied through environment configuration as no key is stored in the notebook or manuscript code.}.
The notebook embeds a 12-record demonstration dataset with explicit \texttt{train}, \texttt{val}, and \texttt{test} splits to keep the method visible. A separate validation experiment tests the end-to-end workflow on non-toy data.
The tutorial code was tested with the package versions reported in \ref{app:execution_env}. GEPA is an active optimisation component in the DSPy ecosystem, and its API or serialisation behaviour may change across package releases. Readers should therefore use the recorded package versions when reproducing the tutorial exactly, and should treat newer versions as requiring re-verification of the notebook, compiled artefact, and evaluation outputs.

\subsection*{Step 1: Define the task contract}

The DSPy signature operationalises the task contract. It states what information the model receives and what fields it must return. In different evidence-synthesis workflows, this contract may describe a screening decision (i.e., \texttt{include} or \texttt{exclude}), an extraction schema, a coding rubric, a risk-of-bias judgement, or a tool-selection policy. It also contains the starting prompt harness, which is a concise but domain-rich description of the task, scope, reasoning or extraction procedure, and guidance for satisfying the fixed output contract. This context engineering step matters most for smaller student models. A thin instruction gives GEPA little material to vary so it can propose a meaningful alternative prompt, while a richer harness gives the reflection model more structure to inspect and work with when examples fail. Useful context includes task scope, near-miss cases, expected evidence use, and the required form of the output.

Listing~\ref{lst:task_contract} mirrors the Colab notebook and gives the screening instantiation used in the worked example. The same pattern can be rewritten for other tasks. For data extraction, the input fields might include the full text and extraction schema, while the output fields might include structured study characteristics, effect sizes, populations, interventions, comparators, outcomes, and evidence spans. For tool selection, the input fields might include the task request and available tools, while the output fields might include the selected tool, arguments, and confidence or fallback policy. 
In the screening instantiation, the inclusion and exclusion criteria are supplied through the \texttt{criteria} input rather than being embedded as mutable prompt text. This separation is deliberate. The criteria represent the scientific decision rule agreed for the review; GEPA should not rewrite/mutate them. GEPA is allowed to revise the prompt harness around those criteria so that the student model applies them more reliably.  Once defined, the criteria are treated as fixed protocol data, being documented and held constant for a calibration run.

For readability, \texttt{STARTING\_PROMPT} is shown as the conceptual prompt harness; in the DSPy implementation below, the same text is embedded in the \texttt{AbstractScreening} signature docstring.
In the validation experiment as well as in the Colab notebook, the review criteria were extracted from the full text of the systematic review \cite{stradowski2023machine} (referred to as Study 41), which is the secondary study identified as reference [41] in the SESR-Eval \cite{huotala2025sesreval} dataset; the frozen criteria block used in the validation comparison is reported in the Appendix Listing~\ref{lst:baseline_criteria}. 

\begin{lstlisting}[language=Python,caption={DSPy task contract for title and abstract screening.},label={lst:task_contract}]
import os
import dspy

STUDENT_MODEL = "openai/gpt-4.1-mini"  # Replace with the model to calibrate.
REFLECTION_MODEL = "openai/gpt-5-mini"  # Use a stronger model for reflection.
SEED = 10  # Any fixed integer is acceptable; report it for reproducibility.

api_key = os.getenv("OPENAI_API_KEY")
student_lm = dspy.LM(STUDENT_MODEL, temperature=0.0, api_key=api_key)
reflection_lm = dspy.LM(
    REFLECTION_MODEL, temperature=1.0, max_tokens=16000, api_key=api_key
)
dspy.configure(lm=student_lm, adapter=dspy.ChatAdapter(use_json_adapter_fallback=False))

# Documentation-only copy of the starting harness.
# The same text is embedded in the Signature docstring below.

STARTING_PROMPT = """
Screen titles and abstracts for a systematic review of machine-learning-based
software defect prediction.

Context:
- The review focuses on primary empirical studies in software engineering.
- Relevant studies predict defects, bugs, faults, or fault-proneness in software artefacts.
- Relevant studies use machine learning and report an evaluation or comparison.
- Exclude reviews, surveys, bug triage, issue severity, duplicate-report detection,
  and studies that do not predict defective software artefacts.

Output:
- checks: brief fragments that connect the decision to the criteria.
- label: exactly one token, include or exclude.
"""

class AbstractScreening(dspy.Signature):
    """
    Screen titles and abstracts for a systematic review of machine-learning-based
    software defect prediction.

    Context:
    - The review focuses on primary empirical studies in software engineering.
    - Relevant studies predict defects, bugs, faults, or fault-proneness in software artefacts.
    - Relevant studies use machine learning and report an evaluation or comparison.
    - Exclude reviews, surveys, bug triage, issue severity, duplicate-report detection,
      and studies that do not predict defective software artefacts.

    Output:
    - checks: brief fragments that connect the decision to the criteria.
    - label: exactly one token, include or exclude.
    """
    criteria: str = dspy.InputField(desc="Inclusion and exclusion criteria.")
    title: str = dspy.InputField(desc="Paper title.")
    abstract: str = dspy.InputField(desc="Paper abstract.")
    checks: str = dspy.OutputField(desc="Short fragments supporting the decision.")
    label: str = dspy.OutputField(desc="Exactly one token: include or exclude.")

class Screener(dspy.Module):
    def __init__(self):
        super().__init__()
        self.classify = dspy.Predict(AbstractScreening)

    def forward(self, criteria: str, title: str, abstract: str):
        return self.classify(criteria=criteria, title=title, abstract=abstract)
\end{lstlisting}

\subsection*{Step 2: Codify the standard}

Similar to machine learning, GEPA optimises against examples and a metric, not against a vague instruction. The example dataset is split into training examples that teach the task, the validation examples which guide GEPA in optimising the prompts, and the held-out test examples provide a small feasibility check.

The metric is task-specific because different evidence-synthesis tasks have different error costs. In the general protocol, the metric defines what the optimiser should reward, what output formats are valid, and how competing errors should be balanced. For data extraction, this may mean combining exact field matches, partial credit for near-correct values, evidence-span checks, and penalties for unsupported or hallucinated values. For risk-of-bias support, it may mean weighting disagreements so that adjacent-category errors are treated as less serious than errors at opposite ends of the rating scale. For tool-selection tasks, it may mean rewarding the correct tool and valid arguments while penalising unsafe or unnecessary tool calls. For screening, the main trade-off is between false exclusions and false inclusions. Listing~\ref{lst:metric} therefore shows one concrete metric for binary screening, where false exclusions are penalised more strongly than false inclusions.

The metric is the most important site of methodological control. A standard accuracy metric tells the optimiser whether an output was correct, but it does not explain how the output failed or how severe that failure is in the downstream workflow. GEPA can use richer feedback: the metric can return both a numeric score and a short natural-language diagnosis. This feedback gives the reflection model information it can convert into better instructions/prompts. For abstract screening, the metric should encode the review policy. Given that missing a relevant study is usually worse than sending an irrelevant study to full-text review, the example metric here (Listing~\ref{lst:metric} shows the compact metric used in the Colab notebook)\footnote{The validation experiments used the expanded version in \ref{app:metric_design}. It adds case-grounded feedback and stricter field-format instructions while preserving the same label-level asymmetric scoring. The supplementary validation code also includes the helper utilities \texttt{normalize\_label()} and \texttt{\_checks\_format\_ok()}. Listing~\ref{lst:metric_helpers} shows them because label normalisation and trace validation are part of the executable evaluation contract.} penalises false exclusions more strongly than false inclusions. It also checks for a minimal justification in the \texttt{checks} field. This constraint discourages prompt revisions that obtain the right label while dropping the audit trail needed for reproducibility.

Metric design is highly tunable. Researchers can adjust the false-positive score, add penalties for malformed output, require criterion-level justifications, or add task-specific constraints such as ``route uncertain records to human review''. The metric should be fixed before running the final optimisation and reported with the compiled artefact. 

\begin{lstlisting}[language=Python,caption={Compact GEPA metric used in the Colab notebook.},label={lst:metric}]
def has_supporting_checks(pred):
    checks = str(getattr(pred, "checks", "")).strip()
    return bool(checks) and len(checks.split()) >= 3

def screening_metric(gold, pred, trace=None, pred_name=None, pred_trace=None):
    expected = gold.label
    observed = str(getattr(pred, "label", "")).strip().lower()

    if observed not in {"include", "exclude"}:
        score = 0.0
        feedback = "The label must be exactly 'include' or 'exclude'."
    elif observed == expected:
        score = 1.0
        feedback = "Correct screening decision."
    elif expected == "include" and observed == "exclude":
        score = 0.0
        feedback = (
            "False negative. Missing relevant studies "
            "is costly in screening."
        )
    else:
        score = 0.4
        feedback = (
            "False positive. Include only when criteria "
            "are clearly satisfied."
        )

    if not has_supporting_checks(pred):
        score = min(score, 0.6)
        feedback = (
            feedback
            + " Also provide brief checks that justify the label."
        )

    if pred_name is not None:
        return dspy.Prediction(score=score, feedback=feedback)
    return score
\end{lstlisting}

\subsubsection*{Designing the GEPA metric as a task-specific utility function}

The compact metric in Listing~\ref{lst:metric} implements the error-cost policy for binary screening. A plain exact-match objective would assign full credit to a correct label and zero to any error. That is too coarse for systematic review workflows, because a false negative removes a potentially relevant study from the pipeline, whereas a false positive mainly increases downstream screening workload. This asymmetry is well established in screening research \cite{matwin2010wss}. Table~\ref{tab:metric_asymmetry} shows the scoring used in this example and how it can be adjusted for similar use cases.

\begin{table}[tbhp]
\centering
\small
\begin{tabular}{llllr}
\toprule
Gold label & Predicted label & Error type & Screening meaning & Score \\
\midrule
\texttt{include} & \texttt{include} & correct & relevant study retained & 1.0 \\
\texttt{exclude} & \texttt{exclude} & correct & irrelevant study removed & 1.0 \\
\texttt{include} & \texttt{exclude} & false negative & relevant study missed & 0.0 \\
\texttt{exclude} & \texttt{include} & false positive & extra full-text workload & 0.4 \\
\bottomrule
\end{tabular}
\caption{Asymmetric scoring for binary screening. The false-positive value (0.4) is a policy choice: it treats a false inclusion as clearly inferior to a correct decision while distinguishing it from a false exclusion, which is the more severe screening failure.}
\label{tab:metric_asymmetry}
\end{table}

The false-positive score of 0.4 is a policy parameter rather than a universal screening constant. Increasing this value makes the optimiser more tolerant of over-inclusion and can shift the calibrated program towards higher recall and higher downstream full-text screening workload. Decreasing it increases pressure for precision, but may raise the risk of false exclusions. Users applying this protocol to a new review should therefore report the chosen value and, where possible, perform a sensitivity analysis over plausible values such as 0.2, 0.4, and 0.6. The present validation uses 0.4 as a moderate partial-credit setting to demonstrate the protocol, not as a recommended default for all evidence-synthesis tasks.

The GEPA metric should be reported as a calibration-time utility function rather than as the sole performance measure. During optimisation, it combines label correctness, asymmetric error costs, output-validity checks, and textual feedback for reflection. After optimisation, performance should still be reported on held-out records using conventional measures such as accuracy, precision, recall, and $F_1$. ~\ref{app:metric_design} provides the component-level details, shows how label normalisation and trace validation were implemented, and discusses how this design extends to multiclass screening.

\subsubsection*{Why GEPA rather than few-shot optimisation?}

\begin{table}[htbp]
\centering
\fontsize{9pt}{10pt}\selectfont
\begin{tabular}{p{0.15\linewidth}p{0.24\linewidth}p{0.1\linewidth}p{0.23\linewidth}p{0.16\linewidth}}
\toprule
\textbf{Approach} & \textbf{What it optimises} & \textbf{Reflection model?} & \textbf{Best suited for} & \textbf{Limitation} \\
\midrule
Manual prompt revision & Human-written instructions & No & Small exploratory tasks & Poor audit trail; low reproducibility \\
Few-shot bootstrapping & Demonstration selection & No or optional & Tasks where examples drive behaviour & Less direct instruction repair \\
GEPA & Instruction harness using metric feedback & Yes & Boundary cases, structured outputs, task-policy feedback & Higher calibration cost \\
Manual reflection loop & Human- or LLM-generated revisions & Optional & Framework-agnostic settings & Less standardised packaging \\
Fine-tuning & Model weights & No at inference & Large labelled datasets & Higher setup cost; less prompt-level inspectability \\
\bottomrule
\end{tabular}
\caption{Comparison of prompt-calibration and model-adaptation strategies relevant to structured evidence-synthesis workflows.}
\label{tab:optimisation_strategies}
\end{table}

GEPA is used here because the protocol requires feedback-rich revision of the prompt harness, not only selection or bootstrapping of few-shot examples. Few-shot optimisers are useful when the main task is to choose demonstrations for inclusion in a prompt. Structured evidence-synthesis tasks often fail at boundary conditions where a model may include a near-miss study, over-apply an exclusion criterion, omit an evidence field, or choose the wrong workflow tool. These failures are better addressed by revising task instructions from case-grounded metric feedback while preserving the fixed task contract and machine-readable interface. Table~\ref{tab:optimisation_strategies} situates GEPA among common alternatives.

\subsection*{Step 3: Compile the program with GEPA}

With the task contract, examples, and metric fixed, the next step is to compile the unoptimised \texttt{Screener}. Listing~\ref{lst:compile_gepa} shows the minimal GEPA compilation call used in the Colab tutorial.

\begin{lstlisting}[language=Python,caption={Bounded GEPA compilation step.},label={lst:compile_gepa}]
student = Screener()

gepa = dspy.GEPA(
    metric=screening_metric,
    reflection_lm=reflection_lm,
    max_full_evals=2,
    reflection_minibatch_size=2,
    seed=SEED,
)

compiled_screener = gepa.compile(student, trainset=trainset, valset=valset)
\end{lstlisting}

The main budget parameter is \texttt{max\_full\_evals}. It controls how many candidate program variants GEPA can evaluate before returning a compiled or optimised program. Larger values give GEPA more opportunity to improve the prompt harness, but it increases the cost and may over-specialise when the validation set is small. The example uses \texttt{max\_full\_evals=2} only to keep the Colab run inexpensive. The validation experiment varies this parameter, reported as \texttt{max\_eval}, to examine the cost-performance trade-off.

\texttt{reflection\_minibatch\_size} controls how many examples are shown to the reflection model when proposing revised instructions. Small minibatches are cheap and expose local errors; larger minibatches can reveal broader failure patterns, but may mix unrelated cases and increase cost. For small tutorials, values of 2 or 3 are sufficient. For real evidence-synthesis tasks, this value should be increased only when enough diverse labelled or reference examples are available.

\begin{table}[hbp]
\centering
\fontsize{10pt}{11pt}\selectfont

\begin{tabular}{p{0.3\linewidth}p{0.2\linewidth}p{0.2\linewidth}p{0.2\linewidth}}
\toprule
Parameter & Controls & Increase when & Risk \\
\midrule
\texttt{max\_full\_evals} & Search budget & Held-out performance remains below the target operating point and validation examples are sufficiently diverse & Over-specialisation \\
\texttt{reflection\_minibatch\_size} & Examples shown to reflector & Failures are diverse & Diluted feedback \\
\texttt{num\_threads} & Parallel evaluation & Runtime too long & Rate limits \\
\texttt{temperature} (reflection) & Revision diversity & Revisions too similar & Unstable runs \\
\bottomrule
\end{tabular}
\caption{Key GEPA parameters and practical guidance for tuning.}
\label{tab:budget_control}
\end{table}
Other parameters are implementation controls. \texttt{num\_threads} affects evaluation speed, \texttt{track\_stats} supports debugging and trace reporting, and \texttt{skip\_perfect\_score} controls whether already-correct examples are omitted from reflection. Table~\ref{tab:budget_control} summarises the parameters readers need to understand first.
After compilation, the revised instruction text can be inspected and saved as part of the compiled DSPy artefact rather than treated as an undocumented manual prompt edit. ~\ref{app:prompt_harness} shows the relationship between the starting harness and a compiled harness.

\subsubsection*{Cost and budget reporting}

\texttt{max\_full\_evals} should be reported as both an optimisation hyperparameter and a cost driver. GEPA cost depends on student-model evaluations, reflection-model calls, validation-set size, minibatch size, maximum reflection tokens, provider pricing, and retry behaviour. When available, users should record model calls, input tokens, output tokens, wall-clock time, and provider cost. When token accounting is unavailable, they should at least report the GEPA budget, validation-set size, reflection minibatch size, model identifiers, temperatures, seeds, and execution date. Calibration cost is usually one-off or occasional, so it should be interpreted relative to the number of future records, reviews, or reruns over which the compiled artefact will be reused. Table~\ref{tab:cost_budget_reporting} reports the budget and model settings available for this study.

\begin{table}[htbp]
\centering
\fontsize{9pt}{10pt}\selectfont
\begin{tabular}{p{0.18\linewidth}p{0.27\linewidth}p{0.25\linewidth}p{0.13\linewidth}p{0.09\linewidth}}
\toprule
\textbf{Setting} & \textbf{Student model} & \textbf{Reflection model} & \textbf{\texttt{max\_full\_evals}} & \textbf{Seeds} \\
\midrule
Tutorial smoke test
& \texttt{openai/gpt-4.1-mini}
& \texttt{openai/gpt-5-mini}
& 1
& 10 \\
Validation ablation
& \texttt{openrouter/qwen/ qwen-2.5-7b-instruct}
& \texttt{openai/ gpt-5-mini}
& 2, 6, 12, 24
& 10, 15, 25, 35, 42 \\
\bottomrule
\end{tabular}
\caption{Budget and model settings for Colab notebook verification and archived validation. The tutorial used \texttt{reflection\_minibatch\_size=2}; the validation ablation used \texttt{reflection\_minibatch\_size=8}. Exact token-level cost was not available from the local execution logs. Future deployments should log model calls, input tokens, output tokens, wall-clock time, provider cost, model identifiers, temperatures, execution date, \texttt{reflection\_minibatch\_size}, and \texttt{max\_full\_evals}.}
\label{tab:cost_budget_reporting}
\end{table}

\subsection*{Step 4: Package and reuse the artefact}

Next, the compiled program should be saved and reloaded before being used in a study, turning the prompt optimisation into a reusable research artefact. Listing~\ref{lst:package_artifact} evaluates the compiled program, saves it, reloads it into a fresh \texttt{Screener}, and verifies that the reloaded artefact produces the same held-out predictions.

\begin{lstlisting}[language=Python,caption={Evaluation and artefact round trip.},label={lst:package_artifact}]
def evaluate(program, examples):
    correct = 0
    predictions = []
    for example in examples:
        pred = program(
            criteria=example.criteria,
            title=example.title,
            abstract=example.abstract,
        )
        observed = str(getattr(pred, "label", "")).strip().lower()
        expected = str(example.label).strip().lower()
        predictions.append((expected, observed))
        correct += int(observed == expected)
    return {
        "accuracy": correct / len(examples),
        "n": len(examples),
        "predictions": predictions,
    }

test_metrics = evaluate(compiled_screener, testset)
compiled_screener.save("compiled_abstract_screener.json")

reloaded_screener = Screener()
reloaded_screener.load("compiled_abstract_screener.json")
reload_metrics = evaluate(reloaded_screener, testset)

assert reload_metrics["predictions"] == test_metrics["predictions"]
\end{lstlisting}

The Colab notebook uses this round-trip check only to verify artefact reuse\footnote{The larger validation at the end of the notebook below evaluates the protocol on SESR-Eval records}. In the JSON state-only save used here, DSPy preserves the compiled program state needed to reload the same program architecture, including the compiled signature or instruction state and any stored predictor state such as demonstrations or model assignments. 
It does not by itself preserve the full research provenance needed to interpret or reproduce the calibration. That provenance must be recorded separately, including the dataset version, split identifiers, metric source, optimisation configuration, model and provider details, package versions, seeds, adapter settings, and evaluation or prediction logs.
DSPy also supports whole-program saving with \texttt{save\_program=True}, but that option uses \texttt{cloudpickle} and still does not replace the need to document the dataset, metric, splits, provider-side model version, optimisation configuration, and evaluation logs.

\subsection*{Transparency and artefact-level inspectability}

A central aim of the protocol is to make prompt calibration inspectable rather than ephemeral. In ordinary prompt engineering, the final prompt is often the only visible object, while failed variants, metric assumptions, data splits, and output-normalisation decisions are often lost. In this workflow, the compiled LLM workflow is treated as a research artefact with provenance. The saved artefact should therefore be accompanied by the fixed task specification, starting signature, optimisation metric, GEPA configuration, random seeds, split identifiers, compiled prompt harness, software environment, and held-out prediction or output logs.
In this workflow, the compiled LLM workflow is treated as a research artefact with provenance. 
This packaging supports two forms of reproducibility. The first is execution-level reuse: the saved program can be reloaded and evaluated again under recorded conditions. The second is methodological transparency: the task specification, metric, prompt harness, traces, and evaluation outputs can be inspected to determine why a calibration was selected and what trade-offs it encodes. The artefact should therefore be treated as an inspectable research object, not as a guarantee of provider-independent deterministic reproduction.

\subsection*{Tuning and troubleshooting guidance}

Users should tune the method in the following order: (1) check data and label consistency; (2) refine the metric so that feedback identifies specific failure types; (3) increase \texttt{max\_full\_evals}; and (4) adjust \texttt{reflection\_minibatch\_size}. Table~\ref{tab:failure_modes} lists common failure modes and first remedies.

\begin{table}[htbp]
\centering
\fontsize{10pt}{11pt}\selectfont
\begin{tabular}{p{0.22\linewidth}p{0.25\linewidth}p{0.45\linewidth}}
\toprule
\textbf{Symptom} & \textbf{Likely cause} & \textbf{First remedy} \\
\midrule
Invalid outputs & Weak output contract & Tighten label field description; add metric penalty for out-of-schema labels \\
Over-selective decisions & Recall pressure too weak; false negatives under-penalised & Strengthen false-negative feedback; add examples where relevant studies are easy to miss \\
Over-permissive decisions & False-positive score too generous; inclusion evidence too weak & Lower false-positive partial credit; require explicit evidence for inclusion criteria \\
Validation rises, test drops & Over-specialisation & Reduce budget; increase validation diversity; simplify the metric \\
Trace missing or malformed & Output trace not rewarded & Cap score for missing checks; require criterion-level fragments \\
\bottomrule
\end{tabular}
\caption{Common failure modes and recommended first remedies.}
\label{tab:failure_modes}
\end{table}

\subsection*{Minimal replication recipe}

The following steps summarise the complete protocol. Each step corresponds to a section above and a code in the companion notebook.

\begin{enumerate}[
    label=\arabic*.,
    leftmargin=4em,
    labelsep=0.6em
]
    \item Install the package versions recorded in ~\ref{app:execution_env}.
    \item Load source records or documents for the target evidence-synthesis task.
    \item Freeze the task contract, such as criteria, rubric, extraction schema, coding manual, or tool policy.
    \item Define a DSPy \texttt{Signature} that declares inputs, outputs, and the starting prompt harness.
    \item Build \texttt{dspy.Example} objects with explicit train/validation/test splits.
    \item Define the task metric with output-validity checks, feedback, and task-specific error weighting.
    \item Run GEPA with a fixed seed and bounded budget, such as \texttt{max\_full\_evals}.
    \item Save the compiled program to a JSON artefact.
    \item Reload the artefact into a fresh program object and verify matching outputs.
    \item Report task-specific held-out metrics, such as accuracy, precision, recall, $F_1$, extraction agreement, field-level match, weighted agreement, or routing accuracy.
\end{enumerate}

\section*{Method validation}

\noindent This validation checks whether the protocol can expose and document calibration trade-offs under a fixed structured task contract. It is not intended to rank models or to make a general performance claim about GEPA. It demonstrates three points: a smaller student model can be calibrated under a transparent protocol, optimisation budget can change the resulting operating point, and the compiled artefact can be evaluated and reused with documented provenance.\footnote{The validation was conducted in a separate experimental workflow rather than in the companion Colab tutorial notebook. The Colab notebook is an executable teaching artefact that demonstrates the code path, compilation, saving, reloading, and adaptation steps on a small example.}

The validation was instantiated as a binary title-and-abstract screening task using Qwen-2.5-7B-instruct as the student model and SESR-Eval Study 41 as the benchmark case. Study 41 contains 1,194 candidate records for the secondary review \textit{Machine learning in software defect prediction: A business-driven systematic mapping study} \cite{stradowski2023machine}, and is one of 24 software-engineering secondary-study screening datasets in SESR-Eval \cite{huotala2025sesreval}. The inclusion and exclusion criteria were extracted from the Study 41 source article, reviewed as a compact task contract, and frozen before calibration. Across five random seeds, stratified train/validation/test splits were generated under the same fixed criteria.

The principal comparator was the structured unoptimised DSPy program, evaluated before calling \texttt{gepa.compile()}. It used the same \texttt{criteria}, \texttt{title}, and \texttt{abstract} inputs, the same \texttt{checks} and \texttt{label} outputs, and the same student model, temperature, adapter configuration, label normalisation, held-out examples, and confusion-matrix evaluation as the GEPA conditions. The paired deltas therefore compare the same structured output contract before and after GEPA prompt-harness calibration. Where a label-only direct prompt is reported, it is only a practical reference for ordinary direct prompting, not the principal comparator for attributing the effect of GEPA calibration. ~\ref{app:structured_baseline} defines the structured baseline.
The validation varied only the GEPA budget. The tested \texttt{max\_full\_evals} values were 2, 6, 12, and 24, reported as \texttt{max\_eval}. The metric, model, data splits, and evaluation procedure were held constant. Performance was computed on the held-out test set using confusion-matrix counts and is reported as mean $\pm$ sample standard deviation across five runs.

\begin{figure}[t]
\centering
\begin{tikzpicture}
\begin{axis}[
    width=0.65\linewidth,
    height=10.5cm,
    xlabel={\texttt{max\_eval}},
    ylabel={Mean score},
    xtick={0,2,6,12,24},
    xticklabels={BL,2,6,12,24},
    ymin=0.72, ymax=0.98,
    legend style={at={(0.5,-0.15)}, anchor=north, legend columns=4, font=\small},
    grid=major,
    grid style={dashed, gray!30},
    every axis plot/.append style={thick, mark size=2.5pt},
]
\addplot+[color=blue, mark=square*, error bars/.cd, y dir=both, y explicit] coordinates {
    (0,0.788) +- (0,0.004) (2,0.776) +- (0,0.038)
    (6,0.786) +- (0,0.045) (12,0.797) +- (0,0.023)
    (24,0.780) +- (0,0.039)
};
\addplot+[color=red, mark=triangle*, error bars/.cd, y dir=both, y explicit] coordinates {
    (0,0.896) +- (0,0.006) (2,0.938) +- (0,0.033)
    (6,0.907) +- (0,0.050) (12,0.905) +- (0,0.030)
    (24,0.855) +- (0,0.050)
};
\addplot+[color=orange, mark=diamond*, error bars/.cd, y dir=both, y explicit] coordinates {
    (0,0.791) +- (0,0.003) (2,0.762) +- (0,0.052)
    (6,0.792) +- (0,0.072) (12,0.799) +- (0,0.043)
    (24,0.811) +- (0,0.071)
};
\addplot+[color=teal, mark=o, error bars/.cd, y dir=both, y explicit] coordinates {
    (0,0.840) +- (0,0.004) (2,0.839) +- (0,0.019)
    (6,0.842) +- (0,0.021) (12,0.848) +- (0,0.011)
    (24,0.829) +- (0,0.021)
};

\legend{Accuracy, Recall, Precision, $F_1$}
\end{axis}
\end{tikzpicture}
\caption{Budget-sensitive operating points across GEPA settings. BL denotes the structured unoptimised baseline. Points show means across five runs; error bars show $\pm 1$ sample standard deviation. The figure shows that increasing GEPA budget did not produce a monotone improvement curve in this validation: low budget favoured recall, a moderate budget gave the best balanced mean performance, and the largest tested budget reduced mean $F_1$ and utility.}
\label{fig:budget_performance}
\end{figure}
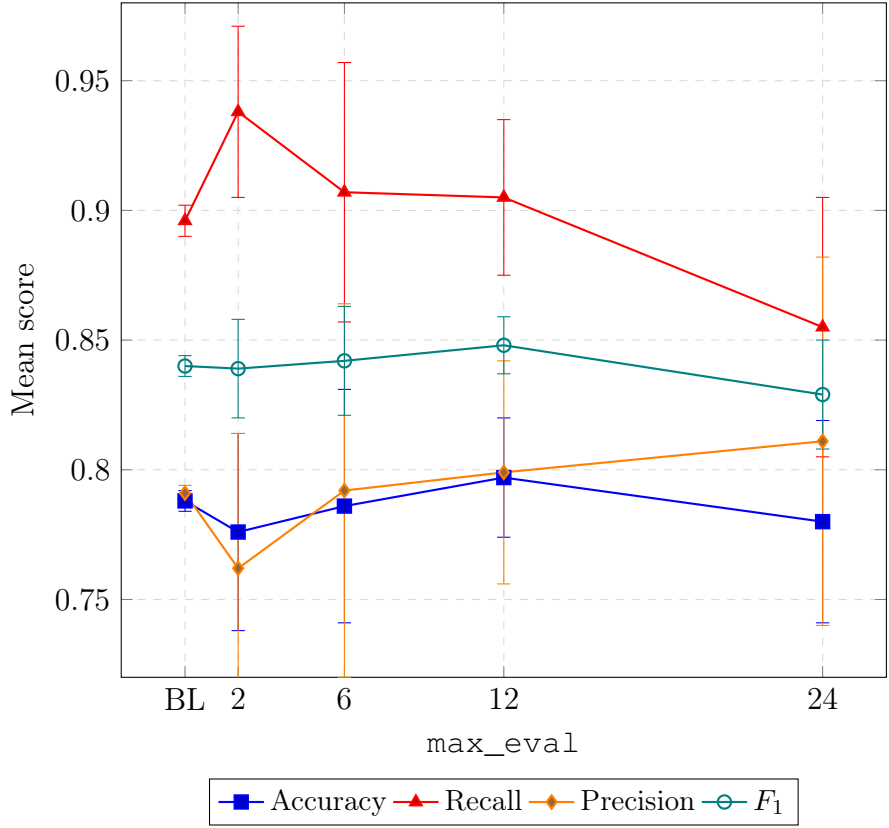

The observed shifts were modest and should be read as operating-point changes in this validation setting, not as evidence of universal optimisation advantage. The structured baseline achieved accuracy $0.788 \pm 0.004$ and $F_1=0.840 \pm 0.004$, with recall $0.896 \pm 0.006$ and precision $0.791 \pm 0.003$ for the \texttt{include} class. A low GEPA budget ($\mathrm{max\_eval}=2$) produced the highest recall ($0.938 \pm 0.033$) but increased the predicted include rate, implying more downstream full-text screening work. The most balanced setting in this validation was $\mathrm{max\_eval}=12$, with accuracy $0.797 \pm 0.023$, $F_1=0.848 \pm 0.011$, and label-level utility $0.855 \pm 0.007$. Relative to the structured baseline, this setting had $\Delta F_1=+0.008 \pm 0.008$ and $\Delta$utility $=+0.008 \pm 0.005$, while changing the predicted include rate by only $+0.002 \pm 0.061$. A larger budget ($\mathrm{max\_eval}=24$) produced lower mean $F_1$ and utility in this validation setting, illustrating why optimisation budget should be treated as a hyperparameter rather than a simple benefit lever. Figure~\ref{fig:budget_performance} visualises these operating-point shifts; full replication settings, metric tables, paired deltas, and SESR-Eval contextual comparisons are provided in ~\ref{app:validation_ablation}.

\section*{Limitations}
\noindent This method calibrates a model to apply a documented task contract; it does not replace expert judgement in evidence synthesis. Borderline eligibility decisions, ambiguous extracted fields, risk-of-bias judgements, and final synthesis decisions should remain under human oversight. Performance depends on the quality and representativeness of the labelled or reference examples, the stability of the selected model provider, and the fidelity of the metric to the task contract. Very small validation sets can also lead GEPA to over-specialise to local examples, especially when \texttt{max\_full\_evals} is increased without more diverse validation data.

The validation is limited to one binary screening task, one benchmark case, one student model, and a small number of repeated runs. The paired GEPA deltas should therefore be read as operating-point changes under a fixed structured contract, not as a general claim that GEPA will improve all evidence-synthesis tasks or all models. Other tasks, including extraction, coding, risk-of-bias support, and tool routing, require task-specific signatures, reference examples, and metrics, although the calibration-and-packaging protocol remains the same.
Exact numerical reproduction is not guaranteed across model-provider changes. API adapters, serving infrastructure, tokenizer behaviour, and provider-side model versions can change, so reproduction may require pinning the model identifier, SDK, adapter versions, package versions, and execution environment recorded in ~\ref{app:execution_env}. 

\section*{Ethics statements}
This study used public benchmark records and published secondary-study materials, and did not involve human participants, private personal data, intervention, or recruitment. Institutional ethics approval was therefore not required.

\vspace{0.5em}
\noindent \textbf{Funding:}\\
\noindent This research did not receive any specific grant from funding agencies in the public, commercial, or not-for-profit sectors.

\section*{Declaration of interests}
\noindent The author declares that he has no known competing financial interests or personal relationships that could have appeared to influence the work reported in this paper.

\section*{Declaration of generative AI and AI-assisted technologies in the manuscript preparation process}

During the preparation of this work, the author used generative AI and AI-assisted tools to support manuscript drafting, language refinement, brainstorming, code checking, and verification of selected implementation details. After using these tools, the author reviewed, edited, and verified the content as needed and takes full responsibility for the final manuscript, code, analyses, and conclusions.

\section*{Supplementary material and/or additional information}
\noindent Supplementary materials include the executable companion Colab tutorial notebook, the local smoke-test script, the saved compiled-program artefact generated by the tutorial, and the small example dataset embedded in the notebook. The larger validation uses SESR-Eval Study 41 records and the Study 41 source article cited above. Table~\ref{tab:reproducibility_checklist} provides a compact checklist of reported reproducibility items and their locations.
The Colab notebook was executed end-to-end in a clean runtime on 24 April 2026 using the package versions reported in Appendix~\ref{app:execution_env}. The saved artefact reloaded successfully and produced the same smoke-test predictions and metrics as the compiled programme in the Colab run.

\begin{table}[h!]
\centering
\small
\begin{tabular}{p{0.40\linewidth}p{0.12\linewidth}p{0.40\linewidth}}
\toprule
Item & Reported? & Location \\
\midrule
Model identifiers (student + reflection) & Yes & Listing 1, \ref{app:execution_env} \\
Random seeds & Yes & Table~\ref{tab:replication_config} \\
Train/validation/test split sizes & Yes & Table~\ref{tab:replication_config} \\
Structured baseline, direct-prompt reference, and frozen criteria & Yes & \ref{app:baseline_setup}, \ref{app:structured_baseline} \\
Metric source code & Yes & Listing 2, \ref{app:metric_design} \\
Execution environment & Yes & \ref{app:execution_env} \\
Saved compiled artefact & Yes & Supplementary material \\
Evaluation metrics reported & Yes & Tables~\ref{tab:qwen_gepa_budget_main}--\ref{tab:qwen_gepa_budget_deltas} \\
\bottomrule
\end{tabular}
\caption{Reproducibility checklist for the validation experiment.}
\label{tab:reproducibility_checklist}
\end{table}


 \bibliographystyle{elsarticle-num}


\appendix
\section{Prompt references and prompt harnesses}
\label{app:prompt_harness}

The code listings in the article omit the full embedded example records, but the prompt text remains part of the method. The label-only direct-prompt reference documents an ordinary manual-instruction comparator, while the prompt harness records the part of the DSPy program that GEPA may revise. The principal validation baseline is the structured unoptimised DSPy program described in ~\ref{app:structured_baseline}. The inclusion and exclusion criteria are not part of the mutable harness; they are supplied as fixed inputs through the \texttt{criteria} field. This prevents the optimiser from changing the review protocol while allowing it to change how the student model interprets and applies that protocol.

This appendix reports the label-only direct-prompt reference, frozen criteria block, structured unoptimised baseline, starting prompt harness, and an excerpt from a compiled harness. These materials allow readers to inspect what was fixed, what was mutable, and what GEPA changed during calibration. This reporting choice matters because two prompt-based LLM workflows may use the same model and nominally the same task criteria, yet differ materially in task framing, output contract, uncertainty policy, and post-processing.

\subsection{Label-only direct-prompt reference and frozen criteria}
\label{app:baseline_setup}

Listing~\ref{lst:baseline_prompt} gives the exact label-only direct-prompt reference, and Listing~\ref{lst:baseline_criteria} gives the frozen criteria block. The prompt requested only a single screening label and did not require a structured \texttt{checks} trace. It is therefore reported as a practical reference for ordinary direct prompting, not as the principal comparator for attributing the effect of GEPA calibration. The main validation comparison uses the structured unoptimised DSPy baseline described in ~\ref{app:structured_baseline}.

\begin{lstlisting}[language=Python,caption={Label-only direct-prompt reference.},label={lst:baseline_prompt}]
BASELINE_PROMPT = """
You are screening research papers by TITLE and ABSTRACT against the
provided INCLUSION/EXCLUSION criteria.

Output MUST be exactly one label:
  - include : should be included for full-text review
  - exclude : should be excluded

If criteria are missing or ambiguous or your confidence is low, then be
more conservative and give preference to including the study.
""".strip()
\end{lstlisting}

\begin{lstlisting}[language=Python,caption={Frozen criteria block used in the validation comparison.},label={lst:baseline_criteria}]
CRITERIA_TEXT = """
INCLUSION CRITERIA:
  - The paper is an empirical study in software engineering
    (field of computer science).
  - The paper is a primary study.
  - The paper is focused on predicting defects in a software system
    using machine learning techniques.
  - The paper evaluates, analyses, or compares prediction methods and
    provides evidence for efficiency.

EXCLUSION CRITERIA:
  - The paper was not a peer-reviewed article, conference proceeding,
    or book chapter.
  - The publication's language was other than English.
  - The same or limited results were already published and included in
    another study.
"""
\end{lstlisting}

\subsection{Structured unoptimised DSPy baseline}
\label{app:structured_baseline}

The structured unoptimised baseline used the same \texttt{AbstractScreening} signature and \texttt{Screener} module as the GEPA-optimised programs, but was evaluated before calling \texttt{gepa.compile()}. It therefore retained the same input fields, output fields, adapter, student model, temperature, criteria text, label normalisation, and held-out evaluation records as the GEPA conditions. This baseline isolates the effect of GEPA optimisation from the effect of using a structured output contract.

Operationally, the structured baseline is the unoptimised DSPy module produced by \texttt{student = Screener()}. The GEPA condition begins from the same module but passes it to \texttt{gepa.compile( student, trainset=trainset, valset=valset)} before held-out evaluation. This pairing keeps the machine-readable interface constant while allowing GEPA to revise the prompt harness during calibration.

\subsection{Starting prompt harness}

Listing~\ref{lst:starting_prompt} shows the mutable starting prompt harness used in the Colab notebook and as the initial GEPA calibration target.

\begin{lstlisting}[language=Python,caption={Starting prompt harness used in the Colab notebook and as the mutable starting point for GEPA.},label={lst:starting_prompt}]
Screen titles and abstracts for a systematic review of machine-learning-based
software defect prediction.

Context:
- The review focuses on primary empirical studies in software engineering.
- Relevant studies predict defects, bugs, faults, or fault-proneness in software artefacts.
- Relevant studies use machine learning and report an evaluation or comparison.
- Exclude reviews, surveys, bug triage, issue severity, duplicate-report detection,
  and studies that do not predict defective software artefacts.

Output:
- checks: brief fragments that connect the decision to the criteria.
- label: exactly one token, include or exclude.
\end{lstlisting}

\subsection{Example of a compiled harness}

The exact compiled text depends on the labelled examples, metric feedback, reflection model, and optimisation budget. A compiled harness often expands the starting prompt by adding failure-specific guidance, clarifying near-miss categories, and giving more explicit instructions for satisfying the fixed output contract. Listing~\ref{lst:compiled_prompt_excerpt} gives an excerpt from an actual optimised prompt generated during validation. Compared with the starting harness, the compiled harness is more explicit about scope boundaries, clause-level checks, conservative inclusion policy, and the DSPy field structure required by the evaluation pipeline.

\begin{lstlisting}[language=Python,caption={Excerpt from an optimised screening harness produced by GEPA.},label={lst:compiled_prompt_excerpt}]
You are screening research papers by TITLE and ABSTRACT against the
provided INCLUSION/EXCLUSION criteria.

Task summary
- Decide whether a paper is relevant for a narrow systematic review that
  collects primary, peer-reviewed empirical studies that build, evaluate,
  or compare machine-learning models for predicting software
  defects/bugs/faults/failures.
- Use only the provided criteria, TITLE, and ABSTRACT.
- Output must use the exact DSPy markers and format described below and
  nothing else.

Required output format (exact tokens and ordering)
- Return exactly three DSPy blocks in this order and no other text:
  [[ ## checks ## ]]
  (2--5 bullet fragments)
  [[ ## label ## ]]
  (exactly one token: include or exclude)
  [[ ## completed ## ]]

Rules for the [[ ## checks ## ]] block
- Provide 2--5 bullets.
- Each bullet is a short fragment, <= 8 words.
- Bullets must name satisfied or violated inclusion or exclusion criteria.
- Do not add prose or extra explanation in this block.

Interpretation and decision policy
- Treat original experiments, validation studies, and measured results
  as evidence of an empirical primary study.
- Include studies predicting defects, fault-proneness, defect density,
  change-proneness, build breakage, or security bugs when machine
  learning is used and evaluation is reported.
- Exclude bug-report severity, priority, assignee, or triage tasks unless
  they are explicitly framed as defect prediction.
- If peer-review status or duplication is unclear, prefer include rather
  than exclude.
\end{lstlisting}

\section{Metric-design notes for GEPA optimisation}
\label{app:metric_design}

The main text presents a compact metric because the article body teaches the protocol rather than reproducing every local experimental guardrail. This appendix records the richer logic behind the local-testing metric and clarifies which parts are essential, optional, or policy-dependent.

\subsection{Why the metric is more than an accuracy check}

In DSPy, an optimiser does not discover a universally good prompt. It discovers a prompt that maximises the declared metric. The metric should therefore be understood as the task utility that GEPA is asked to maximise. In screening, this utility is asymmetric: missing a relevant record is usually more harmful than forwarding an irrelevant one to full-text review. The metric therefore gives correct predictions full credit, false negatives zero credit, and false positives partial credit. This scoring makes the precision--recall trade-off visible to the optimiser rather than hiding it inside a binary right-or-wrong signal.

\subsection{Richer local-testing metric}

Listing~\ref{lst:metric_full} gives the expanded local-testing metric used for validation and smoke testing.

\begin{lstlisting}[language=Python,caption={Expanded local-testing metric with case-grounded feedback.},label={lst:metric_full}]
def gepa_metric(gold, pred, trace=None, pred_name=None, pred_trace=None):
    gold_lab = normalize_label(getattr(gold, "label", None))
    pred_lab = normalize_label(getattr(pred, "label", None))
    checks = (getattr(pred, "checks", "") or "").strip()

    case = (
        f"\n\nCASE\nCRITERIA:\n{getattr(gold, 'criteria', '')[:800]}"
        f"\n\nTITLE:\n{getattr(gold, 'title', '')[:300]}"
        f"\n\nABSTRACT:\n{getattr(gold, 'abstract', '')[:1200]}\n"
    )

    output_contract = (
        "Return outputs using DSPy field markers exactly:\n"
        "[[ ## checks ## ]]\n"
        "(2-5 bullet fragments, each <= 8 words, max 30 words total)\n"
        "[[ ## label ## ]]\n"
        "(exactly one token: include or exclude)\n"
        "[[ ## completed ## ]]\n"
        "Do not add other text.\n"
    )

    if pred_lab not in {"include", "exclude"}:
        score = 0.0
        feedback = (
            f"Invalid label '{pred_lab}'. "
            f"Label must be exactly one of: include or exclude.\n"
            + output_contract
            + case
        )
    elif pred_lab == gold_lab:
        score = 1.0
        feedback = "Correct.\n" + output_contract
    elif gold_lab == "include" and pred_lab == "exclude":
        score = 0.0
        feedback = (
            "False negative. Prefer INCLUDE when uncertain to avoid "
            "missing relevant studies. Identify which criteria clause "
            "supports inclusion.\n"
            + output_contract
            + case
        )
    else:
        score = 0.4
        feedback = (
            "False positive. Only INCLUDE when the criteria are clearly "
            "satisfied. Identify which clause was over-applied.\n"
            + output_contract
            + case
        )

    if not _checks_format_ok(checks):
        score = min(score, 0.6 if score > 0 else score)
        feedback = "checks format invalid or missing.\n" + output_contract + feedback

    if pred_name is not None:
        return dspy.Prediction(score=score, feedback=feedback)
    return score
\end{lstlisting}

\begin{lstlisting}[language=Python,caption={Validation helper utilities used by the expanded metric.},label={lst:metric_helpers}]
def normalize_label(value):
    text = str(value or "").strip().lower()
    if text in {"include", "included"}:
        return "include"
    if text in {"exclude", "excluded"}:
        return "exclude"
    return text

def _checks_format_ok(checks):
    if not checks:
        return False
    lines = [ln.strip() for ln in checks.splitlines() if ln.strip()]
    if not (2 <= len(lines) <= 5):
        return False
    total_words = 0
    for line in lines:
        words = line.lstrip("-* ").strip().split()
        if len(words) > 8:
            return False
        total_words += len(words)
    return total_words <= 30
\end{lstlisting}

\subsection{What each component is doing}

The expanded metric performs five distinct jobs.

\begin{enumerate}[leftmargin=1.4em]
    \item It scores label correctness, but in a cost-sensitive way rather than a symmetric exact-match way.
    \item It distinguishes false negatives from false positives, so GEPA can prefer prompt variants that preserve recall.
    \item It enforces the output schema, because semantically plausible but structurally invalid labels are unusable in an executable DSPy program.
    \item It preserves a bounded \texttt{checks} trace, which acts as a compact criteria checklist rather than unrestricted reasoning.
    \item It returns failure-specific textual feedback grounded in the actual case, which gives the reflection model more useful revision material than a bare scalar score.
\end{enumerate}

The \texttt{case} block is especially important in reflective optimisation. The metric attaches the truncated criteria, title, and abstract to the feedback. This turns a generic message such as ``false negative'' into an example-grounded diagnosis. The reflection model can then revise the prompt against the actual failure mode instead of inferring a change from aggregate performance alone.

\subsection{Why malformed \texttt{checks} are only partially penalised}

The \texttt{checks} field is not the primary prediction target. The primary target is still the final \texttt{label}. The limited penalty on malformed \texttt{checks} therefore expresses a deliberate compromise: semantic correctness remains most important, but the optimiser is discouraged from dropping the structured trace entirely. If the label is already wrong, malformed \texttt{checks} do not materially change the outcome. If the label is correct, however, capping the score prevents GEPA from learning that checklist compliance is optional.

\subsection{Interpreting the score correctly}

The GEPA metric is a calibration-time optimisation utility, not a neutral performance statistic. It can prefer prompts that slightly lower precision if they reduce the risk of missing relevant studies. For this reason, the paper reports conventional held-out metrics separately after optimisation. The optimisation metric defines which prompt variants are favoured during search; the evaluation metrics describe how the resulting program behaves on unseen examples.

The scalar values should be interpreted as a task-utility scale, not as calibrated probabilities. Their ordinal relationship encodes the policy preference: correct outputs should score highest, false exclusions should be worse than false inclusions in high-recall screening, and malformed outputs should be penalised even when the semantic label is correct. The magnitudes also matter because GEPA compares candidate programs using aggregate metric scores. Raising the false-positive score makes the optimiser more tolerant of over-inclusion and may shift the prompt towards recall; lowering it makes the optimiser more conservative and may favour precision at the cost of missed records. The value 0.4 was used as a moderate partial-credit setting, not as a universal constant. Users should treat these values as policy parameters and should report them with the compiled artefact.

Table~\ref{tab:fp-score-checks-cap} summarises how false-positive scores and checks penalties change the optimisation signal.

\begin{table}[ht]
\small
\centering
\begin{tabular}{p{0.2\linewidth} p{0.20\linewidth} p{0.45\linewidth}}
\toprule
\textbf{Parameter} & \textbf{Setting} & \textbf{Interpretation} \\
\midrule
False-positive score & Lower, e.g., 0.1--0.3
& Stricter inclusion, with higher precision pressure. \\

False-positive score & Moderate, e.g., 0.4--0.6
& Balanced high-recall screening. \\

False-positive score & Higher, e.g., 0.7--0.9
& Strong tolerance for over-inclusion, supporting a recall-prioritising workflow. \\

Checks cap & Lower
& Stronger enforcement of the audit trace. \\

Checks cap & Higher
& Label correctness dominates, while trace compliance is weaker. \\
\bottomrule
\end{tabular}
\caption{Interpretation of false-positive score and checks cap settings.}
\label{tab:fp-score-checks-cap}
\end{table}

\subsection{Sensitivity of the false-positive utility parameter}

The false-positive score determines how strongly the optimisation utility tolerates over-inclusion relative to false exclusion. Table~\ref{tab:fp_utility_sensitivity} recomputes the label-level utility from the archived confusion matrices under three plausible false-positive scores. This is a post-hoc illustration only; it does not rerun GEPA under each score. A full sensitivity analysis would repeat optimisation for each candidate utility value because changing the metric can change the selected prompts.

\begin{table}[ht]
\centering
\small
\begin{tabular}{lccc}
\toprule
Condition & FP score 0.2 & FP score 0.4 & FP score 0.6 \\
\midrule
Structured baseline & 0.818 $\pm$ 0.004 & 0.847 $\pm$ 0.004 & 0.877 $\pm$ 0.004 \\
GEPA $\mathrm{max\_eval}=2$ & 0.813 $\pm$ 0.027 & 0.850 $\pm$ 0.017 & 0.887 $\pm$ 0.009 \\
GEPA $\mathrm{max\_eval}=6$ & 0.817 $\pm$ 0.030 & 0.849 $\pm$ 0.016 & 0.880 $\pm$ 0.006 \\
GEPA $\mathrm{max\_eval}=12$ & 0.826 $\pm$ 0.015 & 0.855 $\pm$ 0.007 & 0.884 $\pm$ 0.004 \\
GEPA $\mathrm{max\_eval}=24$ & 0.806 $\pm$ 0.028 & 0.832 $\pm$ 0.019 & 0.858 $\pm$ 0.016 \\
\bottomrule
\end{tabular}
\caption{Post-hoc label-level utility under alternative false-positive scores, computed from the archived confusion matrices. Values are mean $\pm$ sample SD across five runs.}
\label{tab:fp_utility_sensitivity}
\end{table}

\subsection{How this changes in multiclass settings}

The binary asymmetry used here does not transfer directly to multiclass screening. In a multiclass problem, each error is simultaneously a false negative for the true class and a false positive for the predicted class. A more appropriate design is then a cost matrix over class pairs. For example, predicting \texttt{uncertain} for a truly includable record may be less severe than predicting \texttt{exclude}, because an uncertain record can still be routed to human review. Multiclass metrics therefore require class-pair-specific scores and feedback rather than a single false-positive branch and a single false-negative branch.

\subsection{Practical recommendation}

The main article shows the compact metric because it is readable and teaches the core method. For adaptation work, users should expect to iterate on the metric before they iterate on GEPA budget. In most screening applications, metric design is the most direct way to align prompt optimisation with the review workflow, because the optimiser can only learn the screening policy that the metric makes explicit.

\section{Validation configuration and GEPA budget ablation}
\label{app:validation_ablation}

This appendix reports the validation details that support the concise Method validation section, so readers can reproduce the experiment without turning the main article into a benchmark-style results report.

\begin{table}[h!]
\centering
\fontsize{10pt}{11pt}\selectfont
\begin{tabular}{p{0.2\linewidth}p{0.35\linewidth}p{0.37\linewidth}}
\toprule
Component & Value & Why it matters \\
\midrule
Dataset subset & SESR-Eval Study 41 & Defines task population \\
Criteria source & Study 41 full text & Defines screening protocol \\
Criteria extraction & GPT-5 Nano, reviewed/frozen & Separates extraction from optimisation \\
Student model & Qwen-2.5-7B-instruct & Deployable model under calibration \\
Reflection model & GPT-5-mini & Prompt revision model (calibration only) \\
Train/val/test sizes & 59/59/1076 for the structured baseline and most GEPA runs; one $\mathrm{max\_eval}=12$ GEPA run retained $N=1076$ & Defines labelled budget and held-out evaluation set \\
Seeds & 10, 15, 25, 35, 42 & Enables repeated-run replication \\
GEPA budgets tested & 2, 6, 12, 24 & Defines optimisation effort \\
Metric & Asymmetric utility + checks penalty & Defines optimisation objective \\
Student temperature & 0.0 & Controls stochasticity \\
\bottomrule
\end{tabular}
\caption{Minimum replication configuration for the validation experiment.}
\label{tab:replication_config}
\end{table}

\begin{table}[h!]
\centering
\small
\setlength{\tabcolsep}{5pt}
\resizebox{\linewidth}{!}{%
\begin{tabular}{lcccccccc}
\toprule
Method & Accuracy & Precision & Recall & $F_1$ & MCC & $\kappa$ & $\hat{\pi}(\mathrm{include})$ & Utility \\
\midrule
Structured baseline
& 0.788 $\pm$ 0.004
& 0.791 $\pm$ 0.003
& 0.896 $\pm$ 0.006
& 0.840 $\pm$ 0.004
& 0.539 $\pm$ 0.010
& 0.530 $\pm$ 0.009
& 0.704 $\pm$ 0.005
& 0.847 $\pm$ 0.004
\\
GEPA ($\mathrm{max\_eval}=2$)
& 0.776 $\pm$ 0.038
& 0.762 $\pm$ 0.052
& \textbf{0.938 $\pm$ 0.033}
& 0.839 $\pm$ 0.019
& 0.518 $\pm$ 0.081
& 0.482 $\pm$ 0.106
& 0.768 $\pm$ 0.075
& 0.850 $\pm$ 0.017
\\
GEPA ($\mathrm{max\_eval}=6$)
& 0.786 $\pm$ 0.045
& 0.792 $\pm$ 0.072
& 0.907 $\pm$ 0.050
& 0.842 $\pm$ 0.021
& 0.540 $\pm$ 0.099
& 0.516 $\pm$ 0.129
& 0.719 $\pm$ 0.105
& 0.849 $\pm$ 0.016
\\
GEPA ($\mathrm{max\_eval}=12$)
& \textbf{0.797 $\pm$ 0.023}
& 0.799 $\pm$ 0.043
& 0.905 $\pm$ 0.030
& \textbf{0.848 $\pm$ 0.011}
& \textbf{0.562 $\pm$ 0.053}
& \textbf{0.548 $\pm$ 0.066}
& 0.706 $\pm$ 0.060
& \textbf{0.855 $\pm$ 0.007}
\\
GEPA ($\mathrm{max\_eval}=24$)
& 0.780 $\pm$ 0.039
& \textbf{0.811 $\pm$ 0.071}
& 0.855 $\pm$ 0.050
& 0.829 $\pm$ 0.021
& 0.529 $\pm$ 0.102
& 0.519 $\pm$ 0.107
& 0.661 $\pm$ 0.091
& 0.832 $\pm$ 0.019
\\
\bottomrule
\end{tabular}%
}
\caption{Qwen-2.5-7B abstract screening results (mean $\pm$ sample SD over 5 random splits; positive class is \texttt{include}). $\hat{\pi}(\mathrm{include})=(TP+FP)/N$ approximates downstream full-text screening workload. Utility is the label-level asymmetric utility corresponding to the GEPA cost shaping, $U=(TP+TN+0.4\,FP)/N$. It excludes auxiliary penalties for malformed \texttt{checks} or invalid field formatting, which are used during optimisation but are not represented in the confusion matrix. MCC is Matthews correlation coefficient and $\kappa$ is Cohen's kappa computed from the observed and expected agreement under the run-specific marginal rates. One GEPA run under $\mathrm{max\_eval}=12$ evaluated $N=1076$ examples; all other reported structured baseline and GEPA runs used $N=1076$.}
\label{tab:qwen_gepa_budget_main}
\end{table}
\begin{table}[h!]
\centering
\small
\setlength{\tabcolsep}{5pt}
\resizebox{\linewidth}{!}{%
\begin{tabular}{lcccccc}
\toprule
GEPA budget & $\Delta$Accuracy & $\Delta$Precision & $\Delta$Recall & $\Delta F_1$ & $\Delta \hat{\pi}(\mathrm{include})$ & $\Delta$Utility \\
\midrule
$\mathrm{max\_eval}=2$
& $-0.012 \pm 0.036$
& $-0.028 \pm 0.051$
& $+0.042 \pm 0.039$
& $-0.001 \pm 0.018$
& $+0.064 \pm 0.078$
& $+0.003 \pm 0.016$
\\
$\mathrm{max\_eval}=6$
& $-0.002 \pm 0.045$
& $+0.001 \pm 0.072$
& $+0.010 \pm 0.049$
& $+0.002 \pm 0.020$
& $+0.015 \pm 0.106$
& $+0.001 \pm 0.015$
\\
$\mathrm{max\_eval}=12$
& $+0.009 \pm 0.020$
& $+0.009 \pm 0.041$
& $+0.009 \pm 0.034$
& $+0.008 \pm 0.008$
& $+0.002 \pm 0.061$
& $+0.008 \pm 0.005$
\\
$\mathrm{max\_eval}=24$
& $-0.008 \pm 0.039$
& $+0.021 \pm 0.071$
& $-0.041 \pm 0.047$
& $-0.011 \pm 0.020$
& $-0.043 \pm 0.089$
& $-0.015 \pm 0.018$
\\
\bottomrule
\end{tabular}%
}
\caption{Differences (mean $\pm$ sample SD over 5 splits) between each GEPA setting and the structured baseline on the corresponding seed. Positive values indicate higher metric values relative to baseline. The $\mathrm{max\_eval}=12$, seed 10 GEPA run used $N=1076$, while the supplied structured baseline for the same seed used $N=1076$; rerunning or recomputing that baseline on the retained common subset would make the paired comparison exact.}
\label{tab:qwen_gepa_budget_deltas}
\end{table}
\subsection{Interpreting the budget ablation}

Across five repeated splits, the structured baseline achieved an accuracy of $0.788 \pm 0.004$ and an $F_1$ of $0.840 \pm 0.004$, with recall $0.896 \pm 0.006$ and precision $0.791 \pm 0.003$ for the \texttt{include} class. The gold include prevalence was approximately $0.621$ across all splits.
With a small optimisation budget ($\mathrm{max\_eval}=2$), GEPA increases recall to $0.938 \pm 0.033$ while reducing precision to $0.762 \pm 0.052$ and increasing the predicted include rate to $0.768 \pm 0.075$. The paired deltas show that this setting primarily reduces false negatives at the cost of additional false positives. Relative to the structured baseline, recall increases by $+0.042 \pm 0.039$, precision decreases by $-0.028 \pm 0.051$, and the predicted include rate increases by $+0.064 \pm 0.078$. This pattern is consistent with the asymmetric metric shaping that penalises false negatives more strongly than false positives.

In this validation setting, $\mathrm{max\_eval}=12$ produced the best balanced operating point. It had the highest mean $F_1$ ($0.848 \pm 0.011$), accuracy ($0.797 \pm 0.023$), and label-level utility ($0.855 \pm 0.007$), with recall $0.905 \pm 0.030$ and precision $0.799 \pm 0.043$. Relative to the structured baseline on corresponding seeds, $\mathrm{max\_eval}=12$ had $\Delta F_1=+0.008 \pm 0.008$ and $\Delta$utility $=+0.008 \pm 0.005$, while changing the predicted include rate by only $+0.002 \pm 0.061$. MCC and Cohen's $\kappa$ show the same broad pattern, which is consistent with a modest balanced shift rather than only a large shift in predicted include rate.

Increasing the budget to $\mathrm{max\_eval}=24$ produced a non-monotonic outcome. Mean recall ($0.855 \pm 0.050$), $F_1$ ($0.829 \pm 0.021$), and utility ($0.832 \pm 0.019$) fell below the structured baseline, despite higher mean precision ($0.811 \pm 0.071$). This pattern is compatible with instability or over-specialisation under a limited validation signal. It shows why optimisation budget should be treated as a controlled hyperparameter rather than a monotone benefit lever.
For abstract screening workflows that prioritise minimising missed relevant studies, $\mathrm{max\_eval}=2$ offers the highest recall at the cost of workload. In contrast, $\mathrm{max\_eval}=12$ offers the best balanced operating point with almost no mean increase in predicted include rate. The lower mean values at \texttt{max\_eval}=24 are consistent with over-specialisation under a small validation signal. The validation split contains only 59 examples, so repeated reflective revisions may begin to fit idiosyncratic boundary cases rather than the broader held-out distribution. This does not prove that larger GEPA budgets always overfit. It illustrates a practical constraint: the safe optimisation budget is bounded by the size and diversity of the validation set. In small-label settings, users should treat \texttt{max\_full\_evals} as a hyperparameter, inspect held-out performance across seeds, and prefer the smallest budget that gives stable held-out behaviour.

\subsection{Contextual comparison with SESR-Eval}

The SESR-Eval benchmark provides useful scale context, although their testing protocols are not directly comparable. In their testing GPT-4.1 mini tended to produce the best results. In the validation setting here, the optimised Study 41 workflow reached a mean accuracy of $0.833 \pm 0.008$ when optimised with GPT-4.1 mini. SESR-Eval reports an aggregate Study 41 accuracy of $0.78$ across all evaluated LLMs in their Table IX, and an average secondary-study accuracy of $0.73$ for GPT-4.1 mini across 24 studies in Table VII \cite{huotala2025sesreval}. These figures should be interpreted only as indicative because the prompt architecture, output requirements, and experimental setup differ. SESR-Eval used a more elaborate screening prompt and Likert-style decision procedure, whereas the present study evaluates a calibrated binary screening workflow designed to foreground reproducibility and prompt optimisation. Alos, the present validation places the calibrated 7B-class student model in the same broad numerical range as the published SESR-Eval aggregate figures for Study 41. Direct ranking is not appropriate again because the prompt architecture, output format, decision procedure, and evaluation setup differ. This context matters because the SESR-Eval benchmark was dominated by large proprietary or frontier-scale models, including several OpenAI models, Claude 3.7 Sonnet, Llama 4 Maverick (400B), and DeepSeek R1 (671B). Ministral 8B was the only clearly small open model in their comparison set \cite{huotala2025sesreval}.

\section{Execution environment used for local verification}
\label{app:execution_env}

\begin{table}[htbp]
\centering
\fontsize{10pt}{11pt}\selectfont

\begin{tabular}{p{0.36\linewidth}p{0.56\linewidth}}
\toprule
Component & Verified configuration \\
\midrule
Verification date & 24 April 2026 \\
Operating system & Windows 11; not required for the method \\
Alternative execution environment & Google Colab or equivalent Python runtime with the listed packages \\
Python & 3.11.15 \\
DSPy & 3.2.0 \\
GEPA package & 0.0.27 \\
LiteLLM & 1.82.6 \\
OpenAI Python SDK & 2.32.0 \\
Pandas & 3.0.2 \\
scikit-learn & 1.8.0 \\
Provider routing convention & LiteLLM model strings with environment-based API credentials \\
Tutorial / smoke-test student model used in verification & \texttt{openai/gpt-4.1-mini}; replaceable with another supported student model \\
Tutorial / smoke-test reflection model used in verification & \texttt{openai/gpt-5-mini}; replaceable with another stronger reflection model \\
Validation student model in archived experiment & \texttt{openrouter/qwen/qwen-2.5-7b-instruct} \\
Validation reflection model in archived experiment & \texttt{openai/gpt-5-mini} \\
\bottomrule
\end{tabular}
\caption{Compact execution environment record for reproducibility. The listed configuration documents the verified local and archived validation runs; it is not a required platform. The tutorial can be run in Google Colab or another compatible Python environment, and users may substitute supported student and reflection models according to cost, privacy, governance, and availability constraints.}
\label{tab:execution_environment}
\end{table}

Table~\ref{tab:execution_environment} records the environment used for local verification and archived validation, not a required execution environment. The code can run in Google Colab or any compatible Python environment that supports the listed package versions and model-provider configuration. The operating system documents the verified local run only; it is not a methodological dependency. Users may substitute locally hosted or open-weight student models where privacy, cost, or governance requirements favour local inference.

\section{Schematic adaptation for meta-analysis data extraction}
\label{app:meta_analysis_extraction}

This appendix is illustrative only. It is not an empirical validation. It shows how the same calibration pattern can be adapted to a structured extraction task when labelled examples and a scoring metric are available.

In a meta-analysis extraction task, the fixed scientific task contract consists of the review's extraction schema, outcome definitions, eligible effect measures, unit rules, and conventions for prioritising adjusted versus unadjusted estimates. The input may be a full text, a table, or a selected collection of paragraphs. The output is a structured extraction record containing the study identifier, population, intervention or exposure, comparator, outcome, effect measure, numerical estimate, uncertainty measure, sample sizes, and evidence spans. Listing~\ref{lst:meta_analysis_signature} sketches the corresponding DSPy signature.

\begin{lstlisting}[language=Python,caption={Schematic DSPy signature for meta-analysis data extraction.},label={lst:meta_analysis_signature}]
class MetaAnalysisExtraction(dspy.Signature):
    """
    Extract quantitative values needed for meta-analysis from a study report.

    Use only the supplied text.
    Return values only when they are explicitly supported by the text.
    If a value is not reported, return not_reported.
    """

    extraction_schema: str = dspy.InputField(
        desc="Fixed extraction schema, outcome definitions, and unit conventions."
    )
    review_question: str = dspy.InputField(
        desc="Review question and target population/intervention/comparator/outcome."
    )
    source_text: str = dspy.InputField(
        desc="Full text, table text, or selected paragraphs from the study report."
    )

    study_id: str = dspy.OutputField(desc="Study identifier or citation key.")
    population: str = dspy.OutputField(
        desc="Population or sample described in the extracted comparison."
    )
    intervention_or_exposure: str = dspy.OutputField(
        desc="Intervention, exposure, model, or treatment group."
    )
    comparator: str = dspy.OutputField(desc="Comparator or control group.")
    outcome: str = dspy.OutputField(desc="Outcome matching the extraction schema.")
    effect_measure: str = dspy.OutputField(
        desc="Effect measure, such as mean difference, odds ratio, risk ratio, "
             "hazard ratio, correlation, or standardised mean difference."
    )
    effect_value: str = dspy.OutputField(
        desc="Numerical effect estimate or not_reported."
    )
    uncertainty: str = dspy.OutputField(
        desc="Standard error, confidence interval, p value, standard deviation, "
             "or not_reported."
    )
    sample_sizes: str = dspy.OutputField(
        desc="Group-level or total sample sizes needed for analysis, or not_reported."
    )
    evidence_spans: str = dspy.OutputField(
        desc="Short source fragments supporting the extracted values."
    )
    extraction_notes: str = dspy.OutputField(
        desc="Brief notes on ambiguity, conversions, or missing values."
    )
\end{lstlisting}

The extraction metric would differ from the screening metric because the output is a multi-field record rather than a binary label. A compact design is shown in Listing~\ref{lst:extraction_metric_sketch}. It would give full credit for correct numeric values within a pre-specified tolerance, partial credit for correct field identification with missing uncertainty, and strong penalties for hallucinated values not supported by the text. It should also penalise missing evidence spans, handle \texttt{not\_reported} explicitly when the gold record indicates missingness, and apply higher penalties when the wrong outcome or comparison is extracted because those errors can corrupt the downstream meta-analysis.

\begin{lstlisting}[language=Python,caption={Schematic metric design for extraction calibration.},label={lst:extraction_metric_sketch}]
def extraction_metric(gold, pred, trace=None, pred_name=None, pred_trace=None):
    """
    Schematic only.
    Score combines:
    - schema validity
    - exact or tolerance-based numeric match
    - correct effect-measure type
    - correct outcome mapping
    - correct sample-size extraction
    - evidence-span support
    - penalty for hallucinated values
    """
    ...
\end{lstlisting}

The optimiser should not be asked to infer pooled meta-analytic results. It should calibrate extraction of study-level values only. Statistical pooling remains a downstream analysis step.

\end{document}